\documentclass[lettersize,journal]{IEEEtran}
\usepackage{amsmath,amsfonts}
\usepackage{algorithmic}
\usepackage{algorithm}
\usepackage{array}
\usepackage[caption=false,font=normalsize,labelfont=sf,textfont=sf]{subfig}
\usepackage{textcomp}
\usepackage{stfloats}
\usepackage{url}
\usepackage{verbatim}
\usepackage{graphicx}
\usepackage{cite}
\usepackage{booktabs}
\usepackage{xcolor}
\usepackage{pifont}
\usepackage{float}
\newcommand{\cmark}{\ding{51}}  

\begin{document}

\title{UniDA3D: A Unified Domain-Adaptive Framework \\ for Multi-View 3D Object Detection}

\author{
Hongjing Wu$^{\dag}$,~\IEEEmembership{Student Member,~IEEE,}
Cheng Chi$^{\dag}$,~\IEEEmembership{Member,~IEEE,}
Jinlin Wu,~\IEEEmembership{Member,~IEEE}
Yanzhao Su,~\IEEEmembership{Member,~IEEE}
Zhen Lei,~\IEEEmembership{Member,~IEEE}
Wenqi Ren$^{*}$,~\IEEEmembership{Member,~IEEE}

\IEEEcompsocitemizethanks{
\IEEEcompsocthanksitem H.~Wu and W.~Ren are with the Shenzhen Campus of Sun Yat-sen University, Shenzhen, China, and with The State Key Laboratory of Blockchain and Data Security, Zhejiang University, Hangzhou, China. E-mail: wuhj28@mail2.sysu.edu.cn, renwq3@mail.sysu.edu.cn.
\IEEEcompsocthanksitem C.~Chi is with the Beijing Academy of Artificial Intelligence, Beijing, China. E-mail: chengchi6666@gmail.com.
\IEEEcompsocthanksitem J.~Wu and Z.~Lei are with the National Laboratory of Pattern Recognition, Institute of Automation, Chinese Academy of Sciences, Beijing, China. E-mail: 2993376081@qq.com, zhen.lei@ia.ac.cn.
\IEEEcompsocthanksitem Y.~Su is with the Xi'an Research Institute of High-tech, Xi'an, China. E-mail: syzlhh@163.com.
}
\thanks{$^{\dag}$ These authors contributed equally to this work.}
\thanks{$^{*}$ Corresponding author.}
\thanks{Manuscript submitted to IEEE Transactions on Image Processing.}
}

\markboth{IEEE Transactions on Image Processing}%
{Wu \MakeLowercase{\textit{et al.}}: UniDA3D: A Unified Domain-Adaptive Framework for Multi-View 3D Object Detection}

\IEEEpubid{0000--0000/00\$00.00~\copyright~2025 IEEE}

\maketitle

\begin{abstract}
Camera-only 3D object detection is critical for autonomous driving, offering a cost-effective alternative to LiDAR-based methods. In particular, multi-view 3D object detection has emerged as a promising direction due to its balanced trade-off between performance and cost. However, existing methods often suffer significant performance degradation under complex environmental conditions such as nighttime, fog, and rain, primarily due to their reliance on training data collected mostly in ideal conditions. To address this challenge, we propose UniDA3D, a unified domain-adaptive multi-view 3D object detector designed for robust perception under diverse adverse conditions. UniDA3D formulates nighttime, rainy, and foggy scenes as a unified multi-target domain adaptation problem and leverages a novel query-guided domain discrepancy mitigation (QDDM) module to align object features between source and target domains at both batch and global levels via query-centric adversarial and contrastive learning. Furthermore, we introduce a domain-adaptive teacher--student training pipeline with an exponential-moving-average teacher and dynamically updated high-quality pseudo labels to enhance consistency learning and suppress background noise in unlabeled target domains. In contrast to prior approaches that require separate training for each condition, UniDA3D performs a single unified training process across multiple domains, enabling robust all-weather 3D perception. On a synthesized multi-view 3D benchmark constructed by generating nighttime, rainy, and foggy counterparts from nuScenes (nuScenes-Night, nuScenes-Rain, and nuScenes-Haze), UniDA3D consistently outperforms state-of-the-art camera-only multi-view 3D detectors under extreme conditions, achieving substantial gains in mAP and NDS while maintaining real-time inference efficiency.
\end{abstract}


\section{Introduction}
\IEEEPARstart{C}{amera-only} 3D object detection has become a core perception capability for autonomous driving, where reliable 3D scene understanding is required for safe planning and decision-making. Compared with LiDAR-centric solutions, camera-based systems are attractive for large-scale deployment due to their lower hardware cost, easier maintenance, and richer appearance cues. In particular, multi-view 3D object detection, which exploits multiple synchronized cameras around the vehicle, has emerged as a promising paradigm that achieves a favorable trade-off between accuracy and deployment cost. Compared with monocular 3D detectors~\cite{li2024unimode,yan2024monocd,huang2022monodtr} and LiDAR-based 3D detectors~\cite{zhou2020end,chae2024towards,fan2023once}, multi-view camera-only methods provide a better balance between perception performance and deployment cost. Recent multi-view detectors~\cite{li2024bevformer,li2023bevdepth,huang2022bevdet4d,park2022time,wang2023exploring,liu2024ray} have made remarkable progress, with some even approaching or surpassing LiDAR-based methods under standard daytime benchmarks.

Despite this progress, current camera-only 3D detectors are still far from robust in complex real-world environments. In practical deployment, autonomous vehicles must operate across a wide spectrum of adverse conditions, such as nighttime, heavy rain, and dense fog, where illumination, visibility, and image statistics differ drastically from those in clear daytime scenes. However, most existing datasets and models are developed under near-ideal conditions, and naively applying these detectors to extreme weather often leads to severe performance degradation and unstable behavior. This robustness gap is particularly critical for safety-critical perception, where missed detections or inaccurate 3D localization under adverse weather can directly jeopardize downstream planning.
\IEEEpubidadjcol

In this work, we focus on the problem of multi-view 3D object detection under multiple adverse environmental conditions in a \emph{camera-only} setting. Concretely, we consider a training scenario where annotated source data are available only in clear daytime, while the target domains consist of nighttime, rainy, and foggy conditions without ground-truth labels. The goal is to learn a single detector that generalizes well across all these target domains, without training separate models for each weather type.

Existing solutions face several key challenges in this setting. First, the domain shifts induced by different adverse conditions are heterogeneous and multi-faceted: nighttime introduces low-light noise and color shifts, fog causes low contrast and depth-dependent visibility loss, and rain introduces localized streaks and partial occlusions. A generic domain adaptation strategy designed for a single domain pair (e.g., clear $\rightarrow$ night) is often insufficient to simultaneously handle these diverse shifts. Second, multi-view 3D detection involves not only image-level appearance but also cross-view geometric reasoning and depth estimation, making domain discrepancies propagate from the image space to the 3D representation space. Simply aligning 2D feature maps may fail to correct errors accumulated in 3D queries or BEV representations. Third, collecting densely annotated multi-view 3D datasets under extreme conditions is prohibitively expensive, and synthetic image transformations can distort object appearance and geometry, rendering the original annotations suboptimal for direct supervision.

A number of prior works have explored robustness to adverse weather from different angles. Multi-modal approaches~\cite{cai2023objectfusion,liu2022bevfusion,koh2022joint} fuse LiDAR and camera signals to enhance detection in challenging conditions, but they require additional sensors and are thus less suitable for camera-only platforms. In addition, recent multi-modal datasets and fusion frameworks explicitly target robustness in challenging environments by combining LiDAR, radar, RGB, and thermal sensing~\cite{yan2023thermrad,huang2025l4dr,zimmer2023infradet3d}, further demonstrating the benefit of complementary modalities at the cost of increased hardware and system complexity. Domain adaptation methods for 2D detection and LiDAR-based 3D detection~\cite{liu4913864awardistill,hu2023dagl,chen2024cmt,chen2023revisiting,yang2022st3d++}, as well as cross-domain localization under inconsistent environmental conditions~\cite{yin2021i3dloc}, bridge distribution gaps between source and target domains via adversarial learning, knowledge distillation, or pseudo labeling. However, these methods typically target a single target domain or focus on LiDAR or range sensors, and do not explicitly address unified adaptation for camera-based multi-view 3D perception under multiple adverse conditions. Weather-aware modules~\cite{oh2024monowad,li2023monotdp} introduce condition-specific designs or codebooks to improve robustness, yet they usually require dedicated training or parameterization for each weather condition, increasing both computational and maintenance costs.

Moreover, existing perception benchmarks only partially cover our problem setting. Datasets such as Virtual KITTI~\cite{cabon2020virtual,gaidon2016virtual} and SynFog~\cite{xie2024synfog} provide synthetic variations with diverse weather and illumination, but they either focus on LiDAR-based detection or do not provide multi-view camera imagery with 3D annotations across multiple adverse conditions. As a result, there is still a lack of standardized benchmarks for evaluating camera-only multi-view 3D detectors under realistic extreme-weather scenarios, which hinders systematic study of unified domain adaptation in this context.

To address these challenges, we propose \textbf{UniDA3D}, a \textbf{Uni}fied \textbf{D}omain-\textbf{A}daptive framework for multi-view \textbf{3D} object detection that achieves robust all-weather perception through a single training process. UniDA3D formulates nighttime, rainy, and foggy scenarios as a unified multi-target domain adaptation problem, enabling the model to learn shared domain-invariant semantics while preserving condition-specific nuances. A key ingredient of UniDA3D is the \textbf{Query-Guided Domain Discrepancy Mitigation (QDDM)} module, which performs domain alignment at the level of 3D object queries rather than entire feature maps. By leveraging object queries as semantic anchors, QDDM focuses alignment on object-centric features, mitigates background noise, and combines batch-level adversarial training with class-wise global representations to alleviate long-tail category bias across domains. In addition, we adopt a domain-adaptive teacher--student self-training pipeline, where an exponential-moving-average teacher generates dynamically updated high-confidence pseudo labels for unlabeled adverse-weather images, providing reliable high-level supervision tailored to the target distributions.

Beyond the unified adaptation framework, we also construct a multi-view 3D object detection benchmark under extreme conditions, covering rain, fog, and nighttime scenarios. The benchmark consists of multi-view camera streams with 3D annotations and synthesized adverse-weather counterparts, which together support both training and evaluation of camera-only multi-view 3D detectors in challenging environments. Extensive experiments on nuScenes-Night and our adverse-weather benchmark demonstrate that UniDA3D consistently outperforms state-of-the-art camera-only multi-view 3D detectors under extreme conditions, while maintaining competitive performance in clear weather and real-time inference efficiency. Ablation studies further verify the contribution of each component: query-level alignment significantly improves robustness to background shifts, global class memory mitigates long-tail degradation across weather domains, and the teacher--student pipeline improves pseudo-label quality in all target conditions. We also observe that unified training across nighttime, rain, and fog yields better overall performance and parameter efficiency than training separate models for each condition, highlighting the advantage of treating multi-weather adaptation as a single unified problem.

In summary, the main contributions of this work are as follows:
\begin{itemize}
    \item We formulate unified domain adaptation for camera-only multi-view 3D object detection under multiple adverse conditions, and propose UniDA3D, a unified framework that achieves robust all-weather detection through a single training process without training separate models for each weather type.
    \item We design a query-guided domain discrepancy mitigation module that exploits 3D object queries as semantic anchors to align object-level features across domains at both batch and global scales, effectively suppressing background noise and mitigating long-tail category bias.
    \item We develop a domain-adaptive teacher--student self-training pipeline with dynamically updated high-quality pseudo labels, which provides reliable target-domain supervision and stabilizes optimization across different adverse conditions.
    \item We construct a multi-view 3D object detection benchmark under extreme conditions, covering rain, fog, and nighttime; extensive experiments on this benchmark and nuScenes-Night show that UniDA3D delivers state-of-the-art performance under extreme weather while maintaining real-time inference efficiency.
\end{itemize}

\section{Related Work}

\subsection{Perception Dataset for Extreme Conditions}

Achieving dynamic perception of objects in open environments is a major challenge in computer vision, as object appearance can undergo significant changes under complex illumination and weather. Building datasets that explicitly cover diverse conditions is therefore critical for improving robustness. In object tracking, Ye et al.~\cite{ye2022unsupervised} introduced the NAT2021 dataset for nighttime aerial tracking under extreme low-light conditions, while Wu et al.~\cite{wu2025lvptrack} constructed synthetic foggy and nighttime UAV datasets to facilitate domain-adaptive tracking under adverse weather. For 2D scene understanding in driving scenarios, Sakaridis et al.~\cite{sakaridis2018semantic} generated Foggy Cityscapes by applying physically based fog simulation to urban images, and Liu et al.~\cite{liu2022image} built VOC\_Foggy and VOC\_Dark to train and evaluate detectors in foggy and low-light conditions. Beyond these early efforts, ACDC~\cite{sakaridis2021acdc}, Dark Zurich~\cite{sakaridis2019guided}, and BDD100K~\cite{yu2018bdd100k} provide large-scale real-world data with diverse weather and illumination, but mainly focus on single-view 2D perception.
For 3D object detection, the Virtual KITTI series~\cite{cabon2020virtual,gaidon2016virtual} and SynFog~\cite{xie2024synfog} offer photo-realistic synthetic data with varied weather to study robust perception and domain adaptation. Several works~\cite{huang2024sunshine,hahner2021fog,hahner2022lidar} further generate simulated LiDAR data under fog, snow, and rain using physically based models, targeting robustness of LiDAR-based 3D detection in adverse weather. Recently, multi-modal benchmarks that combine LiDAR, radar, RGB, and thermal cameras~\cite{yan2023thermrad,huang2025l4dr,zimmer2023infradet3d} investigate robustness from a multi-sensor perspective, but rely on additional hardware that may not be feasible for camera-only platforms. Overall, most existing resources either target 2D single-view tasks, LiDAR-centric 3D perception, or hardware-rich multi-modal setups, and rarely offer camera-only multi-view imagery with 3D annotations across multiple extreme conditions. Our benchmark is developed to fill this gap by providing multi-view 3D data tailored to nighttime, rain, and fog in a camera-only setting, enabling systematic evaluation of unified domain adaptation for vision-based 3D detection.

\subsection{Multi-view 3D Object Detection}

Multi-view 3D object detection has become a fundamental component of perception systems for autonomous driving, offering a cost-effective alternative to LiDAR by leveraging spatial cues from multiple camera views. Recent methods mainly follow two directions: BEV-based detectors~\cite{huang2021bevdet,li2024bevformer,li2023bevdepth,li2023bevstereo,huang2022bevdet4d} that transform image features into a top-down representation, and sparse query-based detectors~\cite{liu2022petr,park2022time,wang2023exploring,lin2022sparse4d} that directly predict objects from learnable queries. Representative BEV approaches such as BEVDet~\cite{huang2021bevdet}, BEVFormer~\cite{li2024bevformer}, BEVDepth~\cite{li2023bevdepth}, and BEVStereo~\cite{li2023bevstereo} progressively improve temporal modeling and depth estimation, while query-based methods like PETR~\cite{liu2022petr}, SOLOFusion~\cite{park2022time}, StreamPETR~\cite{wang2023exploring}, and Ray Denoising~\cite{liu2024ray} emphasize object-centric temporal reasoning and robust query refinement.
Beyond BEV and query paradigms, unified 3D perception frameworks~\cite{liu2023petrv2,lin2022sparse4d} and stronger backbones or pretraining strategies~\cite{yang2023bevformer,park2021pseudo,li2024unimode,yan2024monocd,huang2022monodtr} further enhance performance, and monocular 3D detectors~\cite{li2024unimode,yan2024monocd,huang2022monodtr} demonstrate that geometric cues can be recovered even from single-view images. However, most existing multi-view detectors are designed and evaluated on benchmarks such as nuScenes~\cite{caesar2020nuscenes}, KITTI~\cite{geiger2013vision}, and Waymo~\cite{sun2020scalability}, which predominantly contain daytime and clear-weather scenes. Illumination shifts, weather-induced visibility degradation, and cross-condition long-tail distributions are seldom modeled explicitly, and explicit domain adaptation mechanisms are rarely incorporated. UniDA3D builds upon a strong multi-view query-based detector and focuses on enhancing its generalization to multiple adverse-weather domains through unified domain-adaptive training.

\subsection{Cross-domain Adaptation}

In many real-world scenarios, significant domain shifts between training and deployment environments pose major challenges to model generalization, making cross-domain adaptation a central research topic. In 2D detection, early works such as domain adaptive Faster R-CNN~\cite{chen2018domain} perform image-level and instance-level alignment, while more recent approaches like DATR~\cite{chen2025datr} employ class-wise and dataset-level alignment for DETR-based detectors~\cite{zhu2020deformable}. IA-YOLO~\cite{liu2022image} explicitly addresses adverse-weather shifts by integrating a differentiable image processing module with adaptive parameter prediction, and benchmarks like BDD100K~\cite{yu2018bdd100k} support systematic evaluation across weather and time-of-day variations. Knowledge distillation and adversarial learning under adverse conditions~\cite{liu4913864awardistill,hu2023dagl} and mean-teacher architectures for UAV tracking~\cite{wu2025lvptrack,ye2022unsupervised} further highlight the effectiveness of teacher–student training for cross-domain robustness.
For 3D object detection, most adaptation efforts~\cite{chen2023revisiting,chen2024cmt,yang2022st3d++,yang2021st3d} focus on LiDAR-based methods and inter-dataset discrepancies. ST3D~\cite{yang2021st3d} and ST3D++~\cite{yang2022st3d++} use self-training with denoised pseudo labels to transfer LiDAR detectors, ReDB~\cite{chen2023revisiting} improves pseudo-label reliability and class balance, and CMT~\cite{chen2024cmt} combines co-training with contrastive learning. Sunshine-to-rainstorm distillation~\cite{huang2024sunshine} and related simulation-based methods~\cite{hahner2021fog,hahner2022lidar} explore cross-weather adaptation for LiDAR-based 3D detection, while MonoWAD~\cite{oh2024monowad} introduces a weather-aware diffusion model for monocular 3D detection. Additional studies~\cite{yin2021i3dloc,yan2023thermrad,huang2025l4dr} investigate localization and detection under inconsistent environmental conditions, often relying on extra modalities such as radar or thermal imaging. However, these approaches typically target a single target domain, focus on LiDAR or monocular settings, or depend on additional sensors, and do not explicitly address unified adaptation for camera-based multi-view 3D perception across multiple adverse-weather domains.

\begin{figure*}[!t]
\centering
\includegraphics[width=\textwidth]{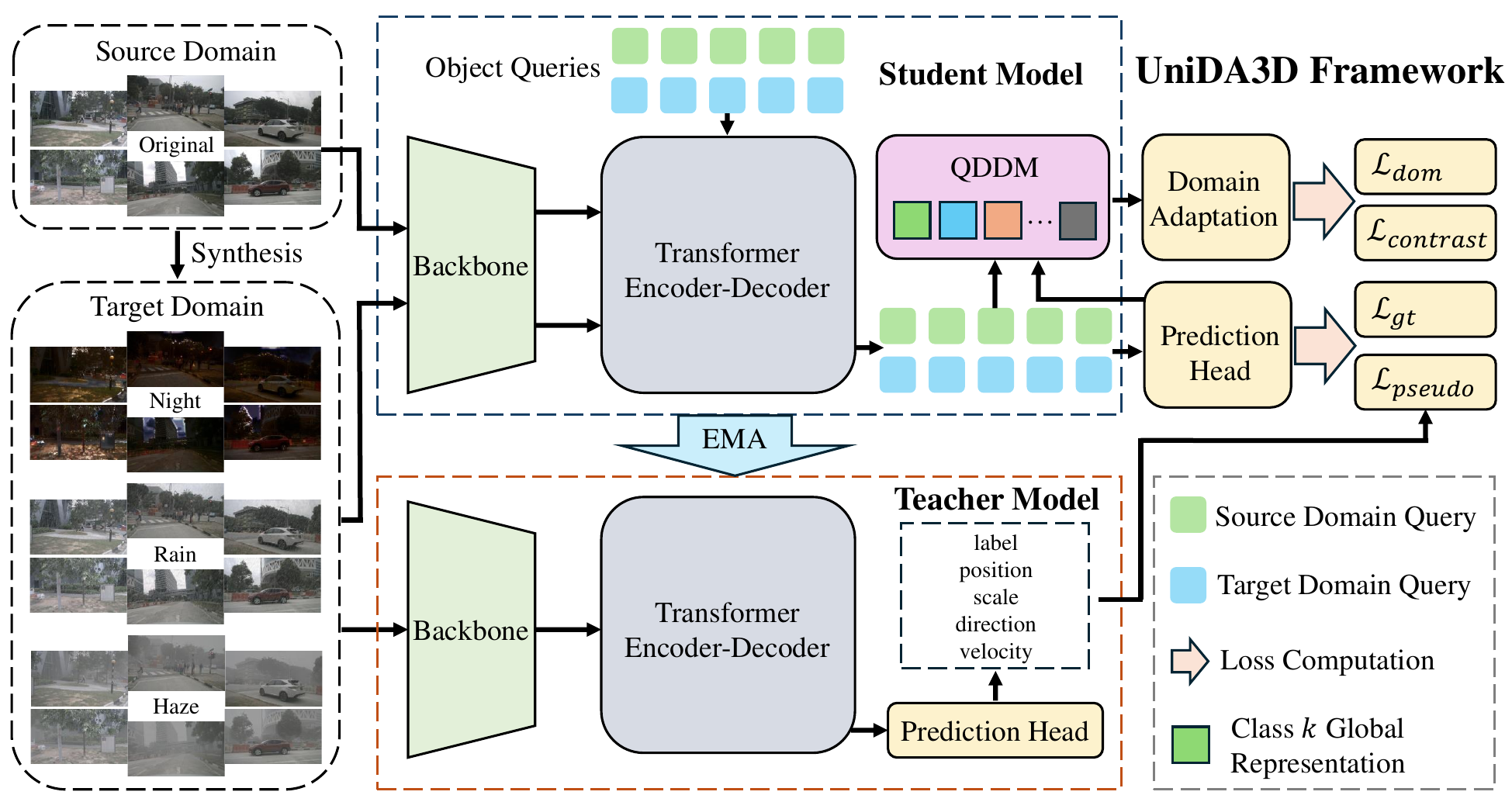}
\caption{Overview of the proposed UniDA3D. UniDA3D first synthesizes nighttime, rainy, and foggy multi-view images to assist training. A domain-adaptive teacher--student self-training pipeline is employed to transfer knowledge from the source domain to the target domains. UniDA3D is further equipped with a query-guided domain discrepancy mitigation (QDDM) module to align object-level features across domains at both batch and global scales.}
\label{fig:framework}
\end{figure*}

\section{Method}

In this section, we illustrate the overall architecture of the proposed UniDA3D in Fig.~\ref{fig:framework}. Our approach adopts a teacher-student self-training pipeline, where the proposed query-guided domain discrepancy mitigation module is integrated into the training process of the student model. UniDA3D first employs different image generators to synthesize multi-view datasets under nighttime, foggy and rainy conditions for model training and evaluation. Subsequently, source-domain images with ground-truth labels and unlabeled target-domain images are jointly fed into the detector network, where domain-consistent learning is conducted under the supervision of multi-domain feature alignment and high-quality pseudo labels. Finally, the 3D detection head outputs the predicted object bounding boxes along with semantic information.

\noindent\textbf{Design motivation.} Extreme-weather images introduce heavy background shifts and long-tail category bias. Instead of aligning whole feature maps, UniDA3D performs query-level alignment centered on object hypotheses to suppress background noise and focus on semantics that transfer across domains. Beyond batch-wise alignment, we maintain class-wise global representations with frequency-adaptive updates to avoid domination by frequent categories. Finally, we unify multiple target domains into a single training procedure with a shared objective, avoiding condition-specific models and enabling one-pass all-weather training. In the following, we describe the three key components of UniDA3D in detail: cross-domain multi-view image synthesis, domain-adaptive self-training, and query-guided domain discrepancy mitigation.

\subsection{Cross-domain Multi-view Image Synthesis}
\label{sec:image_synthesis}

Due to the lack of multi-view 3D detection data under extreme conditions, we synthesize corresponding datasets for nighttime, fog and rain using different image generation methods. 
Specifically, we employ UNIT~\cite{liu2017unsupervised} to synthesize nighttime images from daytime data. The UNIT model is initialized with VGG-128 pre-trained weights. During inference, given a daytime image $I_{\text{day}}$, the corresponding nighttime image $I_{\text{night}}$ is generated as follows:
\begin{equation}
I_{\text{night}} = G_{\text{night}}\left(E_{\text{day}}(I_{\text{day}})\right),
\end{equation}
where $E_{\text{day}}$ denotes the encoder that projects the daytime image into a shared latent space, and $G_{\text{night}}$ denotes the generator that reconstructs a nighttime image from the latent representation.

To simulate foggy weather, we introduce haze effects based on a physical scattering process. Following Koschmieder's law~\cite{he2010single}, the observed hazy image \(H_c(x, y)\) is modeled as:
\begin{equation}
H_c(x, y) = J_c(x, y) \cdot t(x, y) + A_c \cdot (1 - t(x, y)),
\end{equation}
where \((x, y)\) denote spatial coordinates, \(c\) indicates the RGB channels, \(J\) is the original clear image, \(A\) represents the global atmospheric light, and \(t\) captures the fraction of light reaching the camera without scattering. The transmission maps needed for haze rendering are estimated by Depth Anything~\cite{yang2024depth}, enabling realistic fog synthesis over multi-view images.

As for synthesizing rainy images, we adopt a three-stage pipeline~\cite{garg2006photorealistic}, including noise generation, motion blur, and image blending. Given the original image $I_{\text{day}}$, the rainy image $I_{\text{rain}}$ is obtained by:
\begin{equation}
I_{\text{rain}} = \text{Blend}\left(I_{\text{day}}, \text{Blur}\left(\text{Noise}(I_{\text{day}})\right)\right),
\end{equation}
where $\text{Noise}(\cdot)$ generates the initial raindrop noise map, $\text{Blur}(\cdot)$ applies directional motion blur to simulate raindrop streaks, and $\text{Blend}(\cdot, \cdot)$ fuses the clean image and the processed raindrop layer to generate realistic rainy images.

\noindent\textbf{Implementation details of synthesis.} UNIT is implemented at resolution $720{\times}1280$ with identity and cycle-consistency losses weighted by $\lambda_{\mathrm{id}}{=}0.5$ and $\lambda_{\mathrm{cyc}}{=}10$. For haze rendering, we sample the attenuation coefficient $\beta \sim \mathcal{U}(0.6, 1.4)$ and atmospheric light $A \sim \mathcal{U}(0.7, 1.0)$ per image. Depth maps are estimated by Depth Anything v2 (ViT-B) and bilinearly resized to feature scales. For rain, we set streak density $\rho \sim \mathcal{U}(0.02, 0.06)$, length $L \sim \mathcal{U}(7, 15)$ pixels, angle $\theta \sim \mathcal{U}(-15^\circ, 15^\circ)$, motion blur kernel size $k \in [7, 11]$, and blending ratio $\alpha \sim \mathcal{U}(0.15, 0.35)$. We fix random seed to 42 for reproducibility. For each nuScenes sample, all camera views are transformed consistently within a given weather domain (night, haze or rain), while preserving the original calibration and 3D bounding boxes, so that the synthesized images remain geometrically aligned with the existing annotations. During training, these synthesized extreme-condition images are treated as unlabeled target-domain data: we do not use the original nuScenes annotations as direct supervision in the target domains, but rely on teacher-generated pseudo labels instead (see Sec.~\ref{sec:self_training} for details of the teacher--student pipeline).

\subsection{Domain-Adaptive Self-Training}
\label{sec:self_training}

We design a teacher--student pipeline to train the model using data from multiple domains. The model receives the source-domain data $\mathcal{I}_s=\left\{(\mathbf{x}_{i}^{s},\mathbf{y}_{i}^{s})\right\}_{i=1}^{M_s}$ and the target-domain data $\mathcal{I}_{t}^{c}=\left\{\mathbf{x}_{i}^{t}\right\}_{i=1}^{M_t^c}$ as inputs, where $\mathbf{x}$ denotes the images, $\mathbf{y}$ denotes the corresponding annotations, and $i$ indexes training samples. The variable $c$ denotes the type of extreme condition, including nighttime, foggy or rainy scenarios. For each training batch, we randomly select one of the extreme conditions as the target domain to simulate domain variation. Our teacher and student models share an identical 3D object detection architecture, with the teacher model exclusively processing target domain data to generate pseudo labels, thereby addressing the lack of high-level supervision in the target domain. Concretely, the teacher model's predictions undergo Hungarian matching~\cite{kuhn1955hungarian} to assign predictions to objects and filter duplicates, using a matching cost that combines classification confidence and 3D box IoU: $C_{ij} = -\log(p_{ij}) + \lambda_{\text{box}} \cdot (1 - \text{IoU}(b_i, b_j))$, where $p_{ij}$ is the predicted class probability, $b_i$ and $b_j$ are bounding boxes, and $\lambda_{\text{box}}{=}2.0$ balances classification and localization. Predictions are further filtered using a classification confidence threshold $\beta$ to yield reliable pseudo labels $\hat{\mathcal{Y}}_{t}^{c}=\left\{\mathbf{y}_{i}^{t}\right\}_{i=1}^{M_t^c}$. The student model is then trained using both source-domain data with ground-truth annotations and target-domain data with pseudo labels, sharing the same detection loss formulation for both. The teacher model is updated using an Exponential Moving Average (EMA) of the student model's parameters, which promotes stable training and facilitates smooth knowledge transfer, as defined by the following equation:
\begin{equation}
\theta_{\text{teacher}}^{(t)} = \alpha \cdot \theta_{\text{teacher}}^{(t-1)} + (1 - \alpha) \cdot \theta_{\text{student}}^{(t)},
\end{equation}
where $\theta_{\text{teacher}}^{(t)}$ is updated from the previous teacher $\theta_{\text{teacher}}^{(t-1)}$ and the current student $\theta_{\text{student}}^{(t)}$, controlled by a smoothing coefficient $\alpha$.

Pseudo labels are necessary because, although synthetic pipelines can transform clear images into adverse conditions, the transformation may distort object appearance and geometry, making original annotations suboptimal as supervision in the target domain. In realistic deployment, unlabeled target data are far more accessible than high-quality annotations. The teacher model provides high-confidence pseudo labels that are temporally stable (via EMA) and tailored to the target distribution, which improves supervision quality and reduces confirmation bias compared to directly reusing possibly mismatched synthetic labels.

\noindent\textbf{Teacher update and gradient flow.} The teacher network does not participate in back-propagation and is only updated by EMA from the student; teacher forward on target images is performed without gradient tracking and is used solely to generate pseudo labels.

\begin{algorithm}[t]
\caption{Domain-adaptive teacher--student training}
\label{alg:da_ts}
\begin{algorithmic}[1]
\STATE Initialize student $\theta_s$, copy teacher $\theta_t \leftarrow \theta_s$
\FOR{each iteration}
  \STATE Sample a batch: source $(\mathbf{x}^s,\mathbf{y}^s)$ and target $\mathbf{x}^t$
  \STATE Teacher forward on $\mathbf{x}^t$ (no grad) to produce predictions $\hat{\mathbf{y}}^t$
  \STATE Filter pseudo labels by class confidence $\beta$ and NMS
  \STATE Student forward on $\mathbf{x}^s, \mathbf{x}^t$
  \STATE Compute $\mathcal{L}_{\text{gt}}$ on $(\mathbf{x}^s,\mathbf{y}^s)$
  \STATE Compute $\mathcal{L}_{\text{pseudo}}$ on $(\mathbf{x}^t,\hat{\mathbf{y}}^t)$
  \STATE Compute $\mathcal{L}_{\text{dom}}$ and $\mathcal{L}_{\text{contrast}}$ (QDDM)
  \STATE $\mathcal{L}=\mathcal{L}_{\text{gt}}+\mathcal{L}_{\text{pseudo}}+\lambda_{\text{dom}}\mathcal{L}_{\text{dom}}+\lambda_{\text{con}}\mathcal{L}_{\text{contrast}}$
  \STATE Update $\theta_s$ by back-propagation; do not update $\theta_t$ by gradients
  \STATE EMA update teacher: $\theta_t \leftarrow \alpha \theta_t + (1-\alpha)\theta_s$
\ENDFOR
\end{algorithmic}
\end{algorithm}

\noindent\textbf{Schedules.} We linearly ramp up $\lambda_{\text{dom}}$ and $\lambda_{\text{con}}$ from 0 to their maxima ($\lambda_{\text{dom}}{=}0.1$ and $\lambda_{\text{con}}{=}0.1$) in the first 20\% of iterations, and warm up EMA with $\alpha$ increasing from 0.95 to 0.99. We keep a fixed per-class confidence threshold $\beta{=}0.9$, and we detach teacher gradients throughout training.

\noindent\textbf{Loss functions.} For both source and target branches, the detection loss follows the standard formulation in transformer-based 3D detectors~\cite{liu2022petr,wang2023exploring}: $\mathcal{L}_{\text{gt}}$ and $\mathcal{L}_{\text{pseudo}}$ are composed of a classification loss and a 3D bounding-box regression loss (combining L1 and IoU terms) with identical weighting coefficients on source and target samples. This choice ensures that pseudo labels, once filtered by confidence, receive the same type of supervision as ground-truth annotations, while the overall contribution of domain-alignment losses $\mathcal{L}_{\text{dom}}$ and $\mathcal{L}_{\text{contrast}}$ is controlled solely by $\lambda_{\text{dom}}$ and $\lambda_{\text{con}}$.

\begin{figure*}[!t]
\centering
\includegraphics[width=\textwidth]{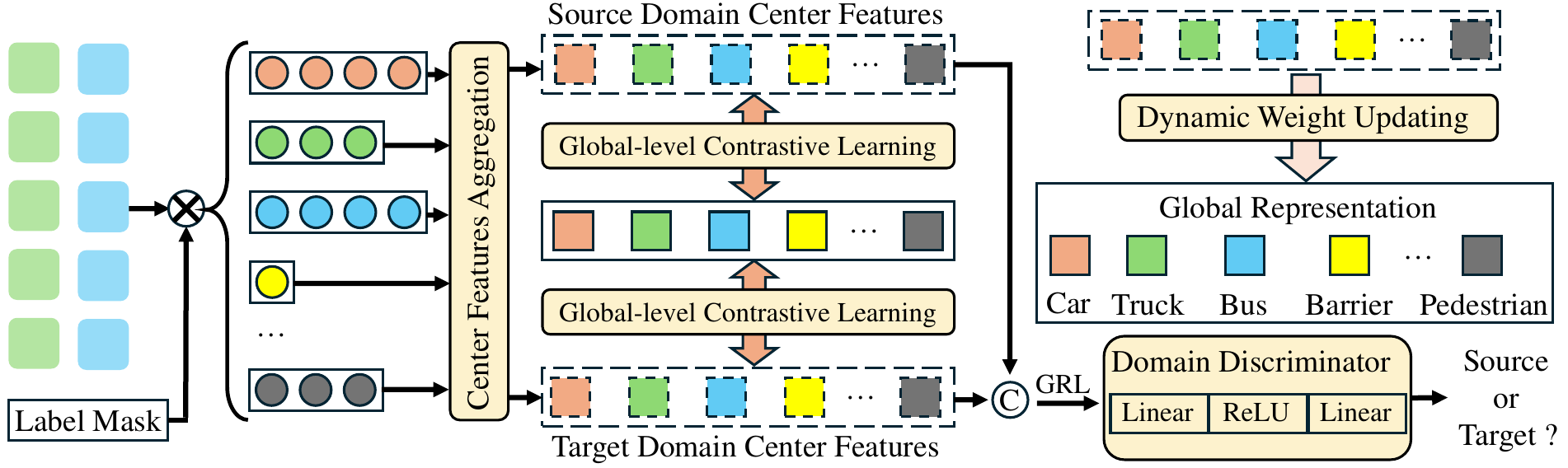}
\caption{Details of the query-guided domain discrepancy mitigation (QDDM) module. QDDM leverages 3D object queries to align source and target features at the object level. It combines adversarial training and contrastive learning with dynamically updated global representation, promoting robust and consistent cross-domain 3D object detection.}
\label{fig:qddm}
\end{figure*}

\subsection{Query-Guided Domain Discrepancy Mitigation module}

\textbf{(\uppercase\expandafter{\romannumeral 1}) Preliminary.}
UniDA3D is developed based on the widely adopted PETR~\cite{liu2022petr} framework. Given a set of multi-camera images 
\(\mathbf{I} = \{ I_i \in \mathbb{R}^{3 \times H_I \times W_I},\; i = 1,2,\dots,N \}\), 
they are first processed by a shared convolutional backbone and a feature pyramid network (FPN) to extract multi-scale 2D image features:
\begin{equation}
\mathbf{F} = \{ F^l_i \in \mathbb{R}^{C^l \times H^l_F \times W^l_F},\; i = 1,\dots,N,\; l = 1,\dots,L \},
\end{equation}
where \(L\) denotes the number of FPN levels, and \(C^l, H^l_F, W^l_F\) are the channel and spatial dimensions of the \(l\)-th level features.
A fixed number of learnable 3D object queries \(\mathbf{Q} \in \mathbb{R}^{N_Q \times D}\) are initialized in the 3D space, where each query \(\mathbf{q}_j \in \mathbb{R}^D\) encodes the spatial location, orientation, and size of a potential object. These queries are associated with a set of 3D reference points \(\mathbf{P} \in \mathbb{R}^{N_Q \times 3}\).
Each 3D reference point \(\mathbf{p}_j \in \mathbb{R}^3\) is projected onto all camera views using the corresponding intrinsic and extrinsic parameters. The resulting 2D projections are used to generate camera-aware positional encodings \(\mathbf{E}_{ij} \in \mathbb{R}^{D}\), capturing the spatial correspondence between 3D queries and 2D image features.
Unlike DETR3D~\cite{wang2022detr3d}, which explicitly samples image features, PETR~\cite{liu2022petr} avoids feature sampling by directly applying cross-attention between 3D object queries and global image features \(\mathbf{F}\), guided by the learned positional encodings \(\mathbf{E}_{ij}\), thereby mitigating information loss caused by projection ambiguity.
Finally, the transformer decoder refines the 3D queries \(\mathbf{Q}\) through multi-layer cross-attention and predicts object class scores and 3D bounding box parameters $(x, y, z, w, h, l, \theta_x, \theta_y)$.

The 3D object queries play a central role in bridging the 3D spatial space and the 2D visual domain, as they serve as the core representation for object-level reasoning across views. Crucially, since each query encodes a potential object hypothesis in the 3D world, it provides a natural anchor for performing object-level domain adaptation. As shown in Fig.~\ref{fig:qddm}, we design query-guided domain discrepancy mitigation module to facilitate robust and transferable 3D perception under cross-domain scenarios.

\textbf{(\uppercase\expandafter{\romannumeral 2}) Batch-level Domain-Consistent Adversarial Learning.}
Compared to reducing domain discrepancies at the full feature map level, employing a query-based domain adversarial training strategy is more effective. This approach mitigates the interference of background noise in domain alignment by focusing solely on minimizing the appearance discrepancies of target objects across domains. Specifically, we retain the 3D object queries as well as their corresponding predicted classification labels of both the source and target domains in the current batch. 
Guided by the predicted class labels, we derive semantic center features for each class through masked feature selection and class-wise feature aggregation. Specifically, we mask out queries with classification confidence below a threshold $\gamma{=}0.5$ to ensure only reliable predictions contribute to class centers. For each query, we first group features across the batch based on class labels, then normalize and aggregate those within the same class to compute the corresponding semantic center features. The process is defined as follows:
\begin{equation}
\mathbf{c}_k^w = \frac{1}{n_k^w} \sum_{i=1}^{N^w} \mathbb{I}(y_i^w = k) \cdot \frac{\mathbf{f}_i^w}{\|\mathbf{f}_i^w\|_2},
\end{equation}
where \( w \in \{s,t\} \) indicates the domain label, with \( s \) and \( t \) representing the source and target domains, respectively. \( \mathbf{f}_i^w \in \mathbb{R}^C \) denotes the feature representation of the \( i \)-th object query, \( y_i^w \in \{1, 2, \dots, K\} \) is its predicted class label, \( \mathbb{I}(\cdot) \) is the indicator function, \( N^w \) is the number of object queries, and \( n_k^w \) denotes the number of instances assigned to class \( k \).

To promote domain-invariant representation learning, we perform adversarial domain alignment on the fused semantic center features. Specifically, the semantic center features from the source and target domains, denoted as $\mathbf{c}_k^s$ and $\mathbf{c}_k^t$ for class $k$, are concatenated and then passed through a Gradient Reversal Layer (GRL), followed by a multi-layer perceptron (MLP) domain discriminator $D_{\text{dom}}$:
\begin{equation}
    \hat{d}_k = D_{\text{dom}}(\mathrm{GRL}(\mathrm{Concat}(\mathbf{c}_k^s, \mathbf{c}_k^t))),
\end{equation}
where $\hat{d}_k$ denotes the predicted domain label. The GRL reverses the gradients during backpropagation, encouraging the semantic center features to be indistinguishable across domains. The domain discriminator is trained using a standard binary cross-entropy loss:
\begin{equation}
    \mathcal{L}_{\text{dom}} = -\sum_k \left[ d_k \log \hat{d}_k + (1 - d_k) \log (1 - \hat{d}_k) \right],
\end{equation}
where $d_k \in \{0, 1\}$ is the domain label, with 0 for source and 1 for target. The three target domains are randomly selected to engage in adversarial training with the source domain's object features, ultimately resulting in a feature extraction network that is invariant to domain discrepancies, thereby enhancing the model's robustness under extreme conditions.

\textbf{(\uppercase\expandafter{\romannumeral 3}) Contrast-Aware Global Representation Guidance.}
Imbalanced category distributions often lead to biased feature representations, particularly when aggregating information across domains. To address this issue, a global representation \( \mathbf{c}_k^{\text{global}} \) is maintained for each class \( k \), serving as a continuously updated semantic reference throughout training. These global embeddings are refined incrementally by incorporating features from incoming batches. Specifically, rather than applying uniform updates, a class-aware weighting strategy is employed, in which the update magnitude is dynamically adjusted according to the relative frequency of each class. The dynamic weight updating allows infrequent classes to contribute more substantially, while moderating the influence of dominant classes. The refinement of \( \mathbf{c}_k^{\text{global}} \) is defined as:
\begin{equation}
\mathbf{c}_k^{\text{global}} \leftarrow \left(1 - \frac{n_k}{n_k + S_k} \right) \cdot \mathbf{c}_k^{\text{global}} + \frac{n_k}{n_k + S_k} \cdot \mathbf{c}_k^{w},
\label{eq:class_adaptive_update}
\end{equation}
where \( S_k \) denotes the accumulated number of samples observed for category \( k \). Consequently, the global representation $\mathbf{c}_k^{\text{global}}$ serves as consistent semantic features for contrastive learning across domains. A class-wise contrastive loss is applied to align batch-level semantic center features \( \mathbf{c}_k^w \) with global representation \( \mathbf{c}^{\text{global}} \), promoting semantic consistency and cross-domain alignment:
\begin{equation}
\mathcal{L}_{\text{contrast}} = 
\sum_{w \in \{s, t\}} \mathcal{L}_{\text{CE}}\left( \mathrm{sim}(\mathbf{c}_k^w,\, \mathbf{c}^{\text{global}}),\; k \right),
\end{equation}
where \( \mathrm{sim}(\cdot, \cdot) \) denotes cosine similarity. The contrastive loss is implemented by first computing temperature-scaled similarities between the batch-level class center $\mathbf{c}_k^w$ and all global class centers: $s_{k,k'} = \exp(\mathrm{sim}(\mathbf{c}_k^w, \mathbf{c}_{k'}^{\text{global}}) / \tau)$ for all classes $k' \in \{1, \dots, K\}$, then applying softmax over all classes to obtain a probability distribution $\mathbf{p}_k = \text{softmax}([s_{k,1}, \dots, s_{k,K}])$, and finally computing cross-entropy against the ground-truth class $k$: $\mathcal{L}_{\text{CE}}(\mathbf{p}_k, k) = -\log(\mathbf{p}_k[k])$. This encourages the batch-level center to be similar to its corresponding global center while being dissimilar to other class centers. Accordingly, the overall training objective can be formulated as:
\begin{equation}
\mathcal{L}_{\text{total}} = \mathcal{L}_{\text{gt}} + \mathcal{L}_{\text{pseudo}} + \lambda_{\text{dom}} \mathcal{L}_{\text{dom}} + \lambda_{\text{con}} \mathcal{L}_{\text{contrast}},
\end{equation}
where \( \mathcal{L}_{\text{gt}} \) denotes the supervised loss computed from ground-truth labels, and \( \mathcal{L}_{\text{pseudo}} \) corresponds to the loss based on pseudo labels. The hyperparameters \( \lambda_{\text{dom}} \) and \( \lambda_{\text{con}} \) balance the contributions of domain alignment and contrastive objectives, with final values set to $\lambda_{\text{dom}}{=}0.1$ and $\lambda_{\text{con}}{=}0.1$ after ramp-up.

\begin{table*}[!t]
\centering
\caption{Comparison with state-of-the-art methods on the nuScenes-Night \texttt{val} set. Models are grouped by backbone (ResNet and V2-99) for a fair comparative analysis. \textbf{Bold} indicates the best result.}
\label{tab:compare1}
\tiny
\resizebox{\textwidth}{!}{
\setlength{\tabcolsep}{4pt}
\begin{tabular}{l|c|c|c|c|c c@{\hspace{1.0\tabcolsep}}c@{\hspace{1.0\tabcolsep}}c@{\hspace{1.0\tabcolsep}}c@{\hspace{1.0\tabcolsep}}} 
\toprule
\textbf{Methods} & \textbf{Backbone} & \textbf{Image Size}  & \textbf{mAP}$\uparrow$  &\textbf{NDS}$\uparrow$  & \textbf{mATE}$\downarrow$ & \textbf{mASE}$\downarrow$   &\textbf{mAOE}$\downarrow$   &\textbf{mAVE}$\downarrow$   &\textbf{mAAE}$\downarrow$ \\
\midrule
BEVDet (CBGS) \cite{huang2021bevdet} & ResNet50 & 256 $\times$ 704 & 6.98 & 14.82 & 91.65 & 50.68 & 84.18 & 136.67 & 60.21 \\
PETR \cite{liu2022petr} & ResNet50 & 512 $\times$ 1408 & 11.97 & 15.51 & 92.69 & 74.40 & 147.76 & 136.79 & 37.72 \\
StreamPETR \cite{wang2023exploring} & ResNet50 & 256 $\times$ 704 & 15.60 & 32.53 & 82.84 & 30.91 & 64.91 & 49.59 & 24.48 \\
SOLOFusion \cite{park2022time} & ResNet50 & 256 $\times$ 704 & 17.84 & 32.91 & 77.64 & 30.78 & 78.28 & 47.02 & 26.41 \\
BEVDet4D \cite{huang2022bevdet4d} & ResNet50 & 256 $\times$ 704 & 22.33 & 37.27 & 76.82 & 28.64 & 61.64 & 49.51 & 22.35 \\
BEVFormer v2 \cite{yang2023bevformer} & ResNet50 & 640 $\times$ 1600 & 25.56 & 35.37 & 80.78 & 29.39 & 53.37 & 89.09 & 21.50 \\
BEVDepth \cite{li2023bevdepth} & ResNet50 & 256 $\times$ 704 & 27.16 & 41.33 & 69.37 & 27.99 & 54.67 & 48.16 & 22.31 \\
BEVStereo \cite{li2023bevstereo} & ResNet50 & 256 $\times$ 704 & 29.37 & 43.35 & \textbf{64.94} & 28.66 & 51.71 & 46.22 & 21.78 \\
BEVFormer-base \cite{li2024bevformer} & R101-DCN & 900 $\times$ 1600 & 30.31 & 43.77 & 74.99 & 29.08 & \textbf{42.68} & 46.74 & 20.34 \\
\textbf{UniDA3D (ours)} & ResNet50 & 256 $\times$ 704 & \textbf{39.74} & \textbf{50.02} & 65.43 & \textbf{27.36} & 49.08 & \textbf{36.80} & \textbf{19.83} \\
\midrule
StreamPETR \cite{wang2023exploring} & V2-99 & 640 $\times$ 1600 & 26.78 & 41.58 & 70.93 & 27.27 & 58.18 & 39.33 & 22.42 \\
PETRv2 \cite{liu2023petrv2} & V2-99 & 320 $\times$ 800 & 28.86 & 31.93 & 84.01 & 71.00 & 154.73 & 48.74 & 21.26 \\
\textbf{UniDA3D (ours)} & V2-99 & 320 $\times$ 800 & \textbf{45.04} & \textbf{53.76} & \textbf{65.87} & \textbf{26.57} & \textbf{43.13} & \textbf{32.58} & \textbf{19.74} \\
\bottomrule
\end{tabular}}
\end{table*}

\begin{table*}[!t]
\centering
\begin{minipage}{0.48\linewidth}
\centering
\caption{Comparison with state-of-the-art methods on the nuScenes-Rain \texttt{val} set.}
\label{tab:compare2}
\scriptsize
\setlength{\tabcolsep}{8pt}
\begin{tabular}{l|c|c|c}
\toprule
\textbf{Methods} & \textbf{Backbone} & \textbf{mAP}$\uparrow$ & \textbf{NDS}$\uparrow$ \\
\midrule
BEVDet (CBGS)~\cite{huang2021bevdet} & ResNet50 & 10.70 & 20.37 \\
PETR~\cite{liu2022petr} & ResNet50 & 15.26 & 17.82 \\
SOLOFusion~\cite{park2022time} & ResNet50 & 17.19 & 33.17 \\
BEVFormer v2~\cite{yang2023bevformer} & ResNet50 & 17.62 & 28.97 \\
BEVFormer-base~\cite{li2024bevformer} & R101-DCN & 17.28 & 31.79 \\
BEVDet4D~\cite{huang2022bevdet4d} & ResNet50 & 23.48 & 37.90 \\
StreamPETR~\cite{wang2023exploring} & ResNet50 & 28.19 & 43.70 \\
BEVDepth~\cite{li2023bevdepth} & ResNet50 & 28.85 & 42.59 \\
BEVStereo~\cite{li2023bevstereo} & ResNet50 & 30.28 & 43.35 \\
\textbf{UniDA3D (ours)} & ResNet50 & \textbf{39.58} & \textbf{49.74} \\
\midrule
PETRv2~\cite{liu2023petrv2} & V2-99 & 32.70 & 35.11 \\
StreamPETR~\cite{wang2023exploring} & V2-99 & 34.26 & 47.53 \\
\textbf{UniDA3D (ours)} & V2-99 & \textbf{45.32} & \textbf{54.71} \\
\bottomrule
\end{tabular}
\end{minipage}
\hfill
\begin{minipage}{0.48\linewidth}
\centering
\caption{Comparison with state-of-the-art methods on the nuScenes-Haze \texttt{val} set.}
\label{tab:compare3}
\scriptsize
\setlength{\tabcolsep}{8pt}
\begin{tabular}{l|c|c|c}
\toprule
\textbf{Methods} & \textbf{Backbone} & \textbf{mAP}$\uparrow$ & \textbf{NDS}$\uparrow$ \\
\midrule
BEVDet (CBGS)~\cite{huang2021bevdet} & ResNet50 & 21.60 & 32.45 \\
PETR~\cite{liu2022petr} & ResNet50 & 24.28 & 23.86 \\
BEVDet4D~\cite{huang2022bevdet4d} & ResNet50 & 28.43 & 42.39 \\
SOLOFusion~\cite{park2022time} & ResNet50 & 31.01 & 43.82 \\
BEVDepth~\cite{li2023bevdepth} & ResNet50 & 31.67 & 45.12 \\
BEVStereo~\cite{li2023bevstereo} & ResNet50 & 33.50 & 46.72 \\
BEVFormer v2~\cite{yang2023bevformer} & ResNet50 & 33.88 & 42.59 \\
StreamPETR~\cite{wang2023exploring} & ResNet50 & 35.50 & 49.38 \\
BEVFormer-base~\cite{li2024bevformer} & R101-DCN & 36.71 & 48.79 \\
\textbf{UniDA3D (ours)} & ResNet50 & \textbf{40.62} & \textbf{50.54} \\
\midrule
PETRv2~\cite{liu2023petrv2} & V2-99 & 36.73 & 37.84 \\
StreamPETR~\cite{wang2023exploring} & V2-99 & 41.50 & 52.56 \\
\textbf{UniDA3D (ours)} & V2-99 & \textbf{46.85} & \textbf{55.90} \\
\bottomrule
\end{tabular}
\end{minipage}
\end{table*}

\section{Experiments}

\subsection{Experimental setup}

\textbf{Datasets.} The nuScenes~\cite{caesar2020nuscenes} dataset is a large-scale autonomous driving dataset collected by Motional, featuring six-camera surround-view images that capture full 360° scenes, along with synchronized LiDAR and RADAR data. It provides comprehensive 3D annotations for tasks such as object detection, tracking, and segmentation. Following the method described in Section~\ref{sec:image_synthesis}, we synthesize three extreme-condition datasets from the original nuScenes data, namely nuScenes-Night, nuScenes-Haze and nuScenes-Rain. The original nuScenes training set contains 28,130 samples, and we synthesize corresponding extreme-condition versions maintaining the same sample count. The validation set contains 6,019 samples, also synthesized to match the original distribution. The training and validation splits remain consistent with the official nuScenes protocol. During unified training, the clear-weather nuScenes training split is treated as the labeled source domain, while all synthesized extreme-condition versions (Night, Haze, Rain) are used as unlabeled target domains; evaluation is always conducted on the synthesized validation splits of each target domain.

\textbf{Metric.} Model performance is evaluated following the nuScenes~\cite{caesar2020nuscenes} protocol, based on mean Average Precision (mAP) and nuScenes Detection Score (NDS). The mAP is computed across multiple distance thresholds over the standard detection classes. NDS aggregates mAP and several True Positive (TP) metrics: mATE (translation error), mASE (scale error), mAOE (orientation error), mAVE (velocity error), and mAAE (attribute error), where lower values indicate better performance in localization, size, orientation, motion, and attribute estimation, respectively. Unless otherwise stated, all reported numbers are obtained on the validation split of each synthesized target domain without using any labels from that domain during training, thereby reflecting the true adaptation capability of the compared methods.

\subsection{Implementation Details}

Our implementation is based on StreamPETR~\cite{wang2023exploring}. We conduct experiments using ResNet-50~\cite{he2016deep} and V2-99~\cite{lee2019energy} as backbone networks, initializing them with pretrained weights from nuImages~\cite{caesar2020nuscenes} and DD3D~\cite{park2021pseudo}, respectively. Input images are resized to $256{\times}704$ for ResNet-50 and $320{\times}800$ for V2-99. UniDA3D is trained with the AdamW optimizer (learning rate $2\times10^{-4}$, weight decay $0.01$), a batch size of 8, and cosine learning rate decay; models with ResNet-50 and V2-99 backbones are trained for 60 and 24 epochs, respectively, following the schedule of StreamPETR.

In each iteration, mini-batches are composed with a 1:1 ratio of source to target samples, and target samples are uniformly drawn from nuScenes-Night, nuScenes-Haze and nuScenes-Rain. We adopt standard data augmentation including color jitter, random resize-and-crop, and horizontal flipping. The EMA coefficient for the teacher model is warmed up from 0.95 to 0.99 as described in Sec.~\ref{sec:self_training}. UniDA3D is developed using Python 3.8 and PyTorch 1.11.0, and is trained and evaluated on four NVIDIA RTX 4090 GPUs with a fixed random seed of 42.

\subsection{State-of-the-art Comparisons}

We compare our proposed UniDA3D with existing state-of-the-art methods on the validation set of our synthesized nuScenes datasets under three extreme conditions. All baselines are trained using the same data composition (clear-weather nuScenes plus synthesized target domains) and training protocol described in the previous subsection to ensure a fair comparison.

\textbf{NuScenes-Night Dataset.} As shown in Table~\ref{tab:compare1}, UniDA3D demonstrates remarkable performance in terms of mAP, NDS, and other metrics. In the V2-99 backbone group, we achieve a mAP of 45.04\% and an NDS of 53.76\%, surpassing the second-best method PETRv2~\cite{liu2023petrv2} by 16.18\% and 21.83\%, respectively. UniDA3D also outperforms StreamPETR~\cite{wang2023exploring} in mATE by a margin of 5.06\%, indicating more accurate localization under low-light conditions where depth cues and object contours are severely degraded. In the ResNet group, we obtain a mAP of 39.74\% and an NDS of 50.02\%, significantly outperforming BEVFormer-base~\cite{li2024bevformer}, which employs a more powerful R101-DCN backbone. This suggests that UniDA3D's unified adaptation strategy yields larger gains than simply scaling up model capacity. Moreover, UniDA3D achieves real-time inference speed, substantially faster than many competing methods; since no extra computation is introduced during inference, the efficiency remains the same as StreamPETR~\cite{wang2023exploring}, confirming that our adaptation modules are training-only.

\textbf{NuScenes-Rain Dataset.} UniDA3D also performs remarkably well in rainy conditions as shown in Table~\ref{tab:compare2}. Using the ResNet-50 backbone, UniDA3D achieves a mAP of 39.58\% and an NDS of 49.74\%, yielding improvements of 9.30\% and 6.39\% over the method BEVStereo~\cite{li2023bevstereo}. With the V2-99 backbone, UniDA3D achieves 45.32\% mAP and 54.71\% NDS, achieving significant improvements over other state-of-the-art methods. The consistent gains across both backbones indicate that UniDA3D is complementary to backbone scaling and can effectively mitigate occlusions and streak-like artifacts introduced by rain, which typically cause detectors to hallucinate or miss small objects.

\textbf{NuScenes-Haze Dataset.} As shown in Table~\ref{tab:compare3}, UniDA3D exhibits strong performance under hazy conditions. Within the V2-99 backbone group, our method achieves a mAP of 46.85\% and an NDS of 55.90\%, outperforming the second-best approach StreamPETR~\cite{wang2023exploring} by 5.35\% and 3.34\%, respectively. Similarly, in the ResNet-50 group, UniDA3D attains a mAP of 40.62\% and an NDS of 50.54\%. Haze mainly reduces contrast and long-range visibility; the improvements in both mAP and NDS reveal that query-level alignment and pseudo-label refinement substantially enhance the stability of 3D localization and orientation estimation in such conditions.

Notably, UniDA3D attains impressive performance with just one unified training run across all extreme conditions, showcasing its all-in-one capability for robust 3D object detection under diverse extreme conditions. Compared with training separate models per weather domain, our unified design avoids overfitting to a single condition and allows complementary regularization between night, rain, and haze, leading to better average performance and deployment simplicity.

\begin{figure}[!t]
\centering
\includegraphics[width=\linewidth]{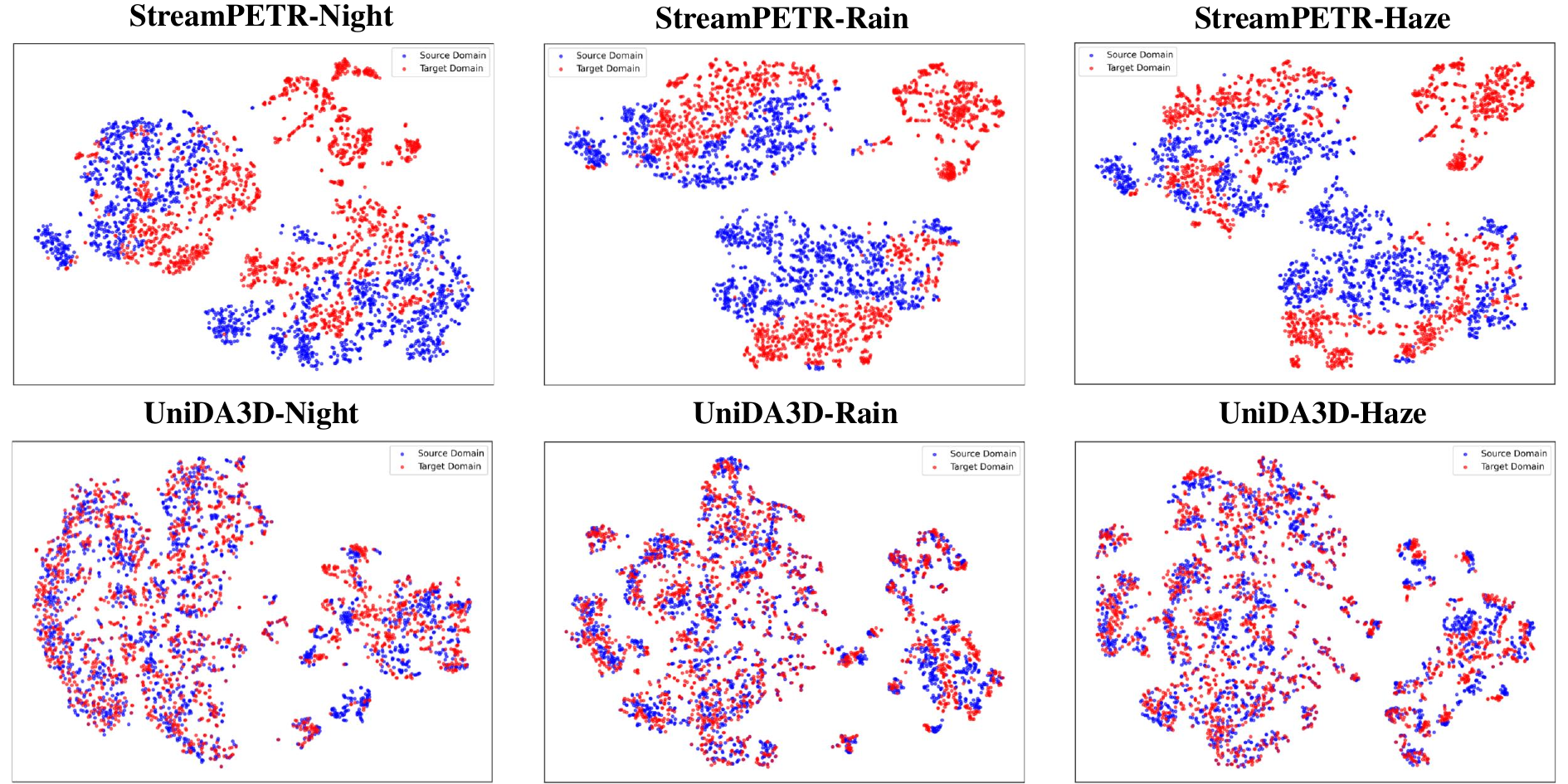}
\caption{t-SNE visualization results (Blue point: Source Domain, Red point: Target Domain). The distribution consistency between red and blue points reflects the discrepancy in features extracted by the model. A smaller gap between the distributions indicates stronger domain adaptation capability.}
\label{fig:tsne}
\end{figure}

\begin{table}[!t]
\scriptsize
\centering
\caption{Effect of EMA ratio. All experiments are conducted on nuScenes-Night \texttt{val} set.}
\label{tab:ema_ratio}
\begin{tabular*}{\columnwidth}{@{\extracolsep{\fill}}cccc@{}}
\toprule
\textbf{EMA Ratio} & \textbf{mAP}$\uparrow$ & \textbf{NDS}$\uparrow$ & \textbf{mATE}$\downarrow$ \\
\midrule
0.9   & 40.83 & 51.10 & 67.08 \\
0.99  & 40.90 & 51.15 & 67.14  \\
0.999 & 33.44 & 45.12 & 65.31 \\
\bottomrule
\end{tabular*}
\end{table}

\begin{table}[!t]
\scriptsize
\centering
\caption{Effect of filtering threshold. The filtering threshold refers to the confidence threshold used to retain high-quality pseudo labels generated by the teacher model. All experiments are conducted on nuScenes-Night \texttt{val} set.}
\label{tab:filtering_threshold}
\begin{tabular*}{\columnwidth}{@{\extracolsep{\fill}}cccc@{}}
\toprule
\textbf{Filtering threshold} & \textbf{mAP}$\uparrow$ & \textbf{NDS}$\uparrow$ & \textbf{mATE}$\downarrow$ \\
\midrule
0.3 & 40.09 & 50.54 & 69.83 \\
0.6 & 40.41 & 51.37 & 66.60 \\
0.9 & 40.90 & 51.15 & 67.14 \\
\bottomrule
\end{tabular*}
\end{table}

\begin{table}[!t]
\scriptsize
\centering
\caption{Effectiveness of Self-training and QDDM modules. All experiments are conducted on nuScenes-Night \texttt{val} set.}
\label{tab:components}
\begin{tabular*}{\columnwidth}{@{\extracolsep{\fill}}ccccc@{}}
\toprule
\textbf{Self-training} & \textbf{QDDM} & \textbf{mAP}$\uparrow$ & \textbf{NDS}$\uparrow$ & \textbf{mATE}$\downarrow$ \\
\midrule
                      &              & 26.78 & 41.58 & 70.93 \\
\cmark                &              & 40.90 & 51.15 & 67.14 \\
                      & \cmark       & 41.88 & 51.72 & 65.91 \\
\cmark                & \cmark       & 45.04 & 53.76 & 65.87 \\
\bottomrule
\end{tabular*}
\end{table}

\subsection{Ablation Study \& Analysis}
\textbf{Study on the training hyper-parameter of UniDA3D.}
We conduct ablation studies on the EMA ratio and filtering threshold. As shown in Table~\ref{tab:ema_ratio}, an EMA ratio of 0.99 achieves the best performance with 40.90\% mAP and 51.15\% NDS. When the EMA ratio is too small (0.9), the teacher updates too quickly and becomes noisy, whereas an excessively large value (0.999) leads to a stale teacher that lags behind the student and accumulates outdated biases, resulting in degraded performance. Table~\ref{tab:filtering_threshold} shows that increasing the filtering threshold consistently improves performance, with the best results achieved at 0.9. This indicates that, under severe domain shifts, it is preferable to rely on fewer but highly confident pseudo labels rather than aggressively exploiting low-confidence predictions, which are more likely to be erroneous. Based on these results, we adopt an EMA ratio of 0.99 and a filtering threshold of 0.9 in our settings.

\textbf{Study on the components of UniDA3D.}
We conduct ablation studies on the two proposed modules in UniDA3D to validate their effectiveness. As shown in Table~\ref{tab:components}, the baseline model (StreamPETR without self-training or QDDM) suffers a significant performance drop in nighttime scenarios due to severe domain discrepancies. Incorporating the teacher--student self-training pipeline significantly enhances the model's domain adaptation capability by exploiting unlabeled target data, and QDDM alone also brings substantial gains by enforcing query-level feature alignment. In particular, the QDDM module reduces the domain gap in object-level feature representations via adversarial and contrastive learning at both batch and global levels, resulting in a mAP of 41.88 (+15.10\%) and an NDS of 51.72 (+10.14\%) over the baseline. When both modules are jointly applied, the model achieves its best robustness, reaching a mAP of 45.04 (+18.26\%) and an NDS of 53.76 (+12.18\%). The additive nature of the gains confirms that UniDA3D benefits from both improved supervision (through pseudo labels) and better feature alignment, rather than relying on only one of these mechanisms.

\textbf{t-SNE Visualization.}
We compare the t-SNE visualization results of feature representations between UniDA3D and StreamPETR under nighttime, rainy and foggy conditions to analyze the domain adaptation effectiveness of our method. As shown in Fig.~\ref{fig:tsne}, StreamPETR exhibits clear distribution gaps between source and target domains across all three extreme conditions, with target features forming clusters that are largely separated from their source counterparts. In contrast, UniDA3D produces more consistent and overlapping feature distributions, demonstrating its ability to extract domain-invariant representations across diverse environments. This qualitative evidence is consistent with the quantitative improvements and suggests that query-level alignment and global class memory jointly encourage weather-robust semantics.

\textbf{Limitations.}
Although UniDA3D achieves impressive results under simulated extreme conditions, there are still two key limitations. Firstly, the target domains are synthesized using image-to-image translation and physically-based rendering methods, which may not fully capture the complexity and diversity of real-world environmental conditions such as sensor artifacts, compound weather patterns, and rare corner cases. Secondly, the robustness of UniDA3D under compounded extreme conditions such as nighttime combined with heavy rain remains to be further investigated, and extending our unified adaptation framework to real-world all-weather datasets is an important direction for future work.

\section{Conclusion}

In this work, we propose UniDA3D, a unified domain-adaptive framework for generalizable 3D object detection under diverse extreme conditions. By integrating a query-guided domain discrepancy mitigation module and a teacher--student self-training pipeline, UniDA3D effectively aligns object-level features across domains and introduces reliable semantic supervision without requiring annotations in target domains. The unified training method eliminates the need for condition-specific models, enabling all-weather 3D perception through a single model.

\bibliographystyle{IEEEtran}
\bibliography{ref}

\end{document}


\title{Supplementary Material for \\ UniDA3D: A Unified Domain-Adaptive Framework for Multi-View 3D Object Detection}

\author{Hongjing Wu, Cheng Chi, Jinlin Wu, Yanzhao Su, Zhen Lei, and Wenqi Ren}

\maketitle

\section{Overview}
This supplementary material provides additional experiments, ablations, and visualizations that complement the main paper. Unless otherwise specified, all notation and experimental settings remain identical to those in the primary manuscript.

\section{Additional Experiments}
We extend the analysis presented in the main submission while following the StreamPETR training pipeline~\cite{wang2023exploring} with the V2-99 backbone~\cite{lee2019energy}. Consistent with the paper, we disable CBGS~\cite{zhu2019class} and reuse the BEVFormer-inspired frame-skipping augmentation~\cite{li2024bevformer}.
\subsection{Speed Testing}
Table~\ref{tab:speed_test} compares the inference speed of UniDA3D with StreamPETR on a single RTX4090 GPU. UniDA3D matches or surpasses the baseline under both ResNet-50 and V2-99 backbones while maintaining robustness in adverse conditions.

\begin{table}[H]
    \centering
    \caption{Inference speed (FPS) comparison on RTX4090.}
    \label{tab:speed_test}
    \begin{tabular}{lcc}
        \toprule
        \textbf{Method} & \textbf{Backbone} & \textbf{FPS} \\
        \midrule
        StreamPETR & ResNet-50 & 28.0 \\
        StreamPETR & V2-99     & 16.7 \\
        UniDA3D (ours) & ResNet-50 & 28.5 \\
        UniDA3D (ours) & V2-99     & 17.9 \\
        \bottomrule
    \end{tabular}
\end{table}

\subsection{Ablation on Target-Domain Training}
We investigate whether training on individual extreme-weather domains differs from hybrid training that mixes all conditions. As summarized in Table~\ref{tab:weather_condition}, hybrid training achieves performance comparable to single-condition training, confirming the multi-domain generalization capability of UniDA3D.

\begin{table}[t]
\centering
\caption{Comparison of training strategies under different weather conditions.}
\small
\label{tab:weather_condition}
\begin{tabular}{c|c|c|c}
\toprule
\textbf{Training Method} & \textbf{mAP} & \textbf{NDS} & \textbf{Condition} \\
\midrule
\multirow{3}{*}{Hybrid} 
  & 45.04 & 53.76 & Night \\
  & 45.32 & 54.71 & Rain \\
  & 46.85 & 55.90 & Haze \\
\midrule
Night & 44.93 & 54.44 & Night \\
Rain  & 45.26 & 54.83 & Rain \\
Haze  & 46.77 & 55.61 & Haze \\
\bottomrule
\end{tabular}
\end{table}

\subsection{Ablation on Loss Weights}
Table~\ref{tab:lambda_ablation} studies the impact of $\lambda_{\text{dom}}$ and $\lambda_{\text{con}}$ on the nuScenes-Night validation set. Setting both weights to 0.1 yields the best trade-off between mAP and NDS.

\begin{table}[t]
\centering
\caption{Ablation on domain adversarial ($\lambda_{\text{dom}}$) and contrastive ($\lambda_{\text{con}}$) loss weights.}
\label{tab:lambda_ablation}
\small
\begin{minipage}{0.48\linewidth}
\centering
\begin{tabular}{c|c|c|c}
\toprule
$\lambda_{\text{dom}}$ & \textbf{mAP}$\uparrow$ & \textbf{NDS}$\uparrow$ & \textbf{mATE}$\downarrow$ \\
\midrule
0.1 & 45.04 & 53.76 & 65.87 \\
0.3 & 42.89 & 53.29 & 65.20 \\
0.6 & 42.91 & 52.86 & 65.96 \\
\bottomrule
\end{tabular}
\vspace{1mm}
\caption*{(a) Effect of $\lambda_{\text{dom}}$.}
\end{minipage}
\hfill
\begin{minipage}{0.48\linewidth}
\centering
\begin{tabular}{c|c|c|c}
\toprule
$\lambda_{\text{con}}$ & \textbf{mAP}$\uparrow$ & \textbf{NDS}$\uparrow$ & \textbf{mATE}$\downarrow$ \\
\midrule
0.1 & 45.04 & 53.76 & 65.87 \\
0.3 & 42.94 & 53.08 & 65.12 \\
0.6 & 41.78 & 52.29 & 64.98 \\
\bottomrule
\end{tabular}
\vspace{1mm}
\caption*{(b) Effect of $\lambda_{\text{con}}$.}
\end{minipage}
\end{table}

\section{Visualization}
Figures~\ref{fig:vis_night}--\ref{fig:vis_haze} showcase detection results under nighttime, rainy, and hazy conditions using bird’s-eye view layouts plus six surround-view cameras. The consistent detection quality across challenging environments underscores the robustness of UniDA3D.

\begin{figure*}[t]
\centering
\suppgraphic{fig/结果可视化/night-2b4441336efa4a649510d33c96ca53aa_pred.png}{Visualization placeholder for nuScenes-Night.}
\caption{Visualization results of UniDA3D on nuScenes-Night.}
\label{fig:vis_night}
\end{figure*}

\begin{figure*}[t]
\centering
\suppgraphic{fig/结果可视化/rain-98592cf269d34c80bdb23c6fa2708db8_pred.png}{Visualization placeholder for nuScenes-Rain.}
\caption{Visualization results of UniDA3D on nuScenes-Rain.}
\label{fig:vis_rain}
\end{figure*}

\begin{figure*}[t]
\centering
\suppgraphic{fig/结果可视化/haze-1f1982e0b2a64ec2b2aba3a41175ab6e_pred.png}{Visualization placeholder for nuScenes-Haze.}
\caption{Visualization results of UniDA3D on nuScenes-Haze.}
\label{fig:vis_haze}
\end{figure*}

\section{Conclusion}
This supplementary document enriches the main paper with speed analyses, additional ablations, and qualitative evidence. Together, these results further validate the effectiveness and practicality of UniDA3D for unified multi-view 3D object detection under extreme conditions.

\bibliographystyle{IEEEtran}
\bibliography{ref}